\documentclass[11pt]{article}
\usepackage{EACL2023}

\usepackage{times}
\usepackage{latexsym}
\usepackage{graphicx}
\usepackage[T1]{fontenc}

\usepackage[utf8]{inputenc}

\usepackage{microtype}

\usepackage{inconsolata}

\title{\Large{\textbf{mahaNLP: A Marathi Natural Language Processing Library}}}

\author{Vidula Magdum\textsuperscript{1,3}, Omkar Dhekane\textsuperscript{1,3}, Sharayu Hiwarkhedkar\textsuperscript{1,3},  Saloni Mittal\textsuperscript{1,3} \\
\textbf{Raviraj Joshi}\textsuperscript{2,3} \\
  \textsuperscript{1} Pune Institute of Computer Technology, Pune, Maharashtra India \\ 
  \textsuperscript{2} Indian Institute of Technology Madras, Chennai, Tamil Nadu India\\ 
  \textsuperscript{3} L3Cube Pune\\
  \texttt{\{vidulamagdum12, omkarjd1212, hiwarkhedkarsharayu, salonimittal12\}@gmail.com} \\
  \texttt{ravirajoshi@gmail.com}}
  
\begin{document}
\maketitle
\begin{abstract}
\small{We present mahaNLP, an open-source natural language processing (NLP) library specifically built for the Marathi language. It aims to enhance the support for the low-resource Indian language Marathi in the field of NLP. It is an easy-to-use, extensible, and modular toolkit for Marathi text analysis built on state-of-the-art MahaBERT-based transformer models. Our work holds significant importance as other existing Indic NLP libraries provide basic Marathi processing support and rely on older models with restricted performance. Our toolkit stands out by offering a comprehensive array of NLP tasks, encompassing both fundamental preprocessing tasks and advanced NLP tasks like sentiment analysis, NER, hate speech detection, and sentence completion. This paper focuses on an overview of the mahaNLP framework, its features, and its usage. This work is a part of the L3Cube MahaNLP initiative, more information about it can be found at \url{https://github.com/l3cube-pune/MarathiNLP}.}
\end{abstract}


\section{Introduction}

\normalsize{Natural Language Processing (NLP) is a major subset of artificial intelligence. It helps to clear up linguistic ambiguity and gives the data a useful quantitative structure for numerous downstream tasks. While NLP has the potential to have a huge impact on the ML community, recent models have primarily focused on English and 6 other languages\footnote{\url{https://datasaur.ai/blog-posts/nlp-work-in-different-languages}} with a significantly high amount of resources \cite{joshi2020state}. There are around 7,000 languages spoken worldwide\footnote{\url{https://en.wikipedia.org/wiki/Lists_of_languages}}. Out of which, approximately 22+ languages are existing Indian languages\footnote{\url{https://en.wikipedia.org/wiki/Languages_of_India}} that are widely spoken not only in India but also throughout the world. Developing models that work for these languages is important for a variety of reasons, including bridging the existing language divide and promoting exploration and research for non-English languages in the growing NLP field\footnote{\url{https://ruder.io/state-of-multilingual-ai/index.html}}.

\begin{figure}[]
  \centering
\includegraphics
[width=\linewidth]
{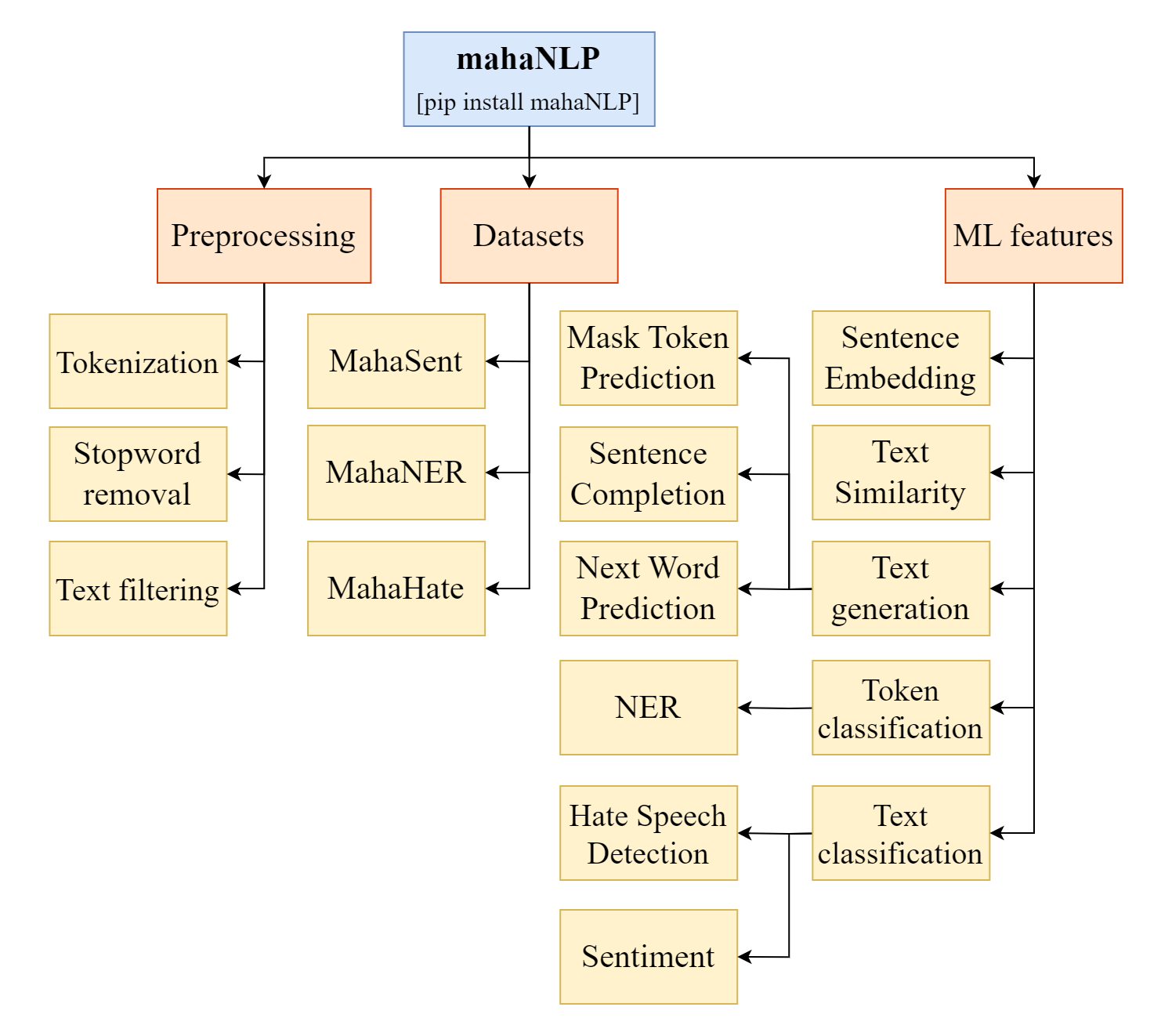}
\caption{A brief overview of the features in mahaNLP library}
\label{fig:overview}
\end{figure}

\begin{figure*}[]
  \centering
\includegraphics[width=\textwidth]{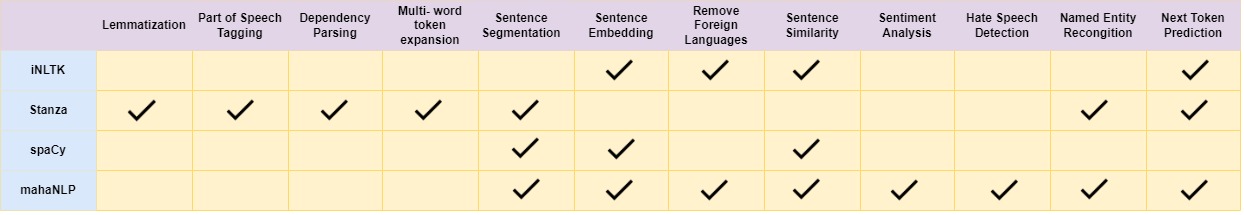}
\caption{A comparison of a few existing libraries and mahaNLP specific to the Marathi language}
\label{fig:comparison}
\end{figure*}

The Marathi language is 11th in the list of popular languages across the globe\footnote{\url{https://en.wikipedia.org/wiki/Marathi_language}}. Despite being a widely spoken language, Marathi-specific NLP monolingual resources are still limited in comparison to other natural languages \cite{joshi2022l3cube_2}. There are many popular open-source tools like Spacy\footnote{\url{https://spacy.io/}}, Stanza\footnote{\url{https://stanfordnlp.github.io/stanza/}}, iNLTK\footnote{\url{https://inltk.readthedocs.io/en/latest/}}, IndicNLP\footnote{\url{https://indicnlp.ai4bharat.org/pages/home/}}, providing features to create effective multilingual NLP systems. 
Though these libraries do support the multilingual functionality for Marathi, they are not exhaustive with respect to the functionalities they support and have their own set of limitations and discrepancies, such as the usage of older architectures to implement multilingual models \cite{ruder2021xtreme}. Basic features like Marathi sentiment analysis, named entity recognition, and hate speech detection is missing in almost all the current multilingual libraries.

The L3Cube-MahaNLP\footnote{\url{https://github.com/l3cube-pune/MarathiNLP}} \cite{joshi2022l3cube} initiative is an umbrella for various tools, datasets, and models that greatly assist in Marathi language processing. With this work, we aim to provide a broader set of functionalities to developers. In this paper, we propose mahaNLP library - a python-based NLP toolkit focused on the Indian language Marathi and majorly built on top of Hugging Face transformer models\footnote{\url{https://huggingface.co/l3cube-pune}}. 
Along with basic language processing features, the open-source toolkit wraps the state-of-the-art monolingual Marathi transformer models for text analysis. 
It encompasses a myriad of features like sentiment and hate analysis, named entity recognition, and a variety of other Marathi language processing features as shown in Figure \ref{fig:overview}. Thus, the mahaNLP library aims to make Marathi NLP more accessible. The demonstration video\footnote{\url{https://youtu.be/12AWux0AtFA?si=dzcO3DrdykmV1HLy}} and example colab\footnote{\href{https://colab.research.google.com/drive/1POx3Bi1cML6-s3Z3u8g8VpqzpoYCyv2q?usp=sharing}{Example colab}} are shared publicly.}




\section{Related Work}


The communities of NLP and machine learning have a prolonged history of developing open-source tools and libraries. There are numerous user-friendly, all-purpose NLP libraries available. NLTK \cite{bird2006nltk}, Stanford
CoreNLP \cite{manning2014stanford}, Spacy \cite{honnibal2017spacy}, AllenNLP \cite{gardner2018allennlp}, Flair \cite{akbik2019flair}, Stanza \cite{qi2020stanza}, Hugging Face Transformers \cite{wolf2019huggingface} and iNLTK \cite{arora2020inltk} are some of the libraries that are primarily concerned with NLP tasks. These libraries provide NLP tasks to the ML community like tokenization, sentence encoding, text normalization, translation, and so on.

However, while some of these libraries do support Marathi, they have a very limited range of NLP tasks and pre-trained language models. Figure~\ref{fig:comparison} illustrates the provisional comparison between mahaNLP and the existing libraries for the language Marathi. The iNLTK library is not actively supported, it is based on very limited datasets and LSTM-based ULMFiT models. Marathi models for features like sentiment and NER are unavailable in SpaCy. MahaNLP attempts to address these issues by offering a broader set of pre-trained language models, including BERT \cite{devlin2018bert}, RoBERTa \cite{liu2019roberta}, and AlBERT \cite{lan2019albert}, as well as a broader set of NLP tasks such as sentiment analysis, hate speech analysis, NER Tagger, and MLM-based modules. The NER model supported by Stanza is based on our L3Cube-MahaNER corpus \cite{joshi2022l3cube_ner}\footnote{\url{https://stanfordnlp.github.io/stanza/ner_models.html}}.


\section{System Design and Architecture}
The \textit{ease-of-access} is an important characteristic of mahaNLP and the system is designed from the user's perspective. These perspectives can be mainly defined as - a \textit{Standard Flow} and a \textit{Model Flow}.

\subsection{Standard Flow}
The \textit{Standard Flow} can be used by a basic programmer who has the least knowledge of the machine learning domain. In this flow, the complex model arguments are isolated from the users. They can use a feature without knowledge of the models used in the background. This flow has constructs similar to standard NLP libraries. The intuitive nature of this flow makes mahaNLP more user-friendly and easily accessible. The pre-processing, tokenizer, datasets, and ML-based models are part of this user flow.




\subsubsection{Pre-process Module}
An initial step in any NLP task is the preprocessing of data. The transformer models in NLP perform much better on cleaned data than raw, unprocessed corpus. The preprocess module helps to provide functions such as the removal of URLs, stopwords, and non-Devanagari words for cleaning Marathi textual data. 


\subsubsection{Tokenizer Module}
A very important step in many NLP tasks is the tokenization of text. The mahaNLP Tokenizer module provides functionalities for sentence-level tokenization (splitting into multiple sentences) and word-level tokenization (splitting into multiple words). 


\subsubsection{Datasets Module}
The mahaNLP library currently supports 3 datasets - MahaHate \cite{joshi2022l3cube_hate} for hate speech detection, MahaSent \cite{joshi2022l3cube_sent}\cite{pingle2023l3cubemahasentmd} for sentiment analysis, and MahaNER \cite{joshi2022l3cube_ner} for named entity recognition. These datasets can mainly be used for finetuning tasks. Each of these corpora can be separately loaded in the pandas data frame. The datasets are cached locally to avoid repeated downloads.

\begin{table}[hbt!]
\centering
\def\arraystretch{1.5}
\scalebox{0.7}{
\begin{tabular}{|p{2.5cm}|p{3.5cm}|p{3.5cm}|}
\hline
\textbf{Feature name} & \textbf{Labels / Outputs} & \textbf{Function name(s)}\\
\hline
Sentiment & positive, negative, neutral & get\_polarity\_score \\
\hline
Hate & hate, non-hate & get\_hate\_score \\
\hline
Tagger & Person (NEP), Location (NEL), Organization (NEO), Measure (NEM), Time (NETI), Date (NED), Designation (ED) & get\_token\_labels, get\_tokens \\
\hline
Auto Complete & next (N) words & next\_word, complete\_sentence \\
\hline
Mask Fill & masked word & predict\_mask \\
\hline
Similarity & similarity score (0-1) & embed\_sentences, get\_similarity\_score \\
\hline
\end{tabular}
}
\caption{Standard Flow ML features}
\label{tab:standard-flow-features}
\end{table}

\subsubsection{Machine learning-based modules}
A set of modules utilize machine learning models in the background to provide the desired functionality. Currently, these features or modules are implemented using state-of-the-art Marathi Transformer models and are described below. The basic syntax to use them is:
\includegraphics[width=\linewidth]{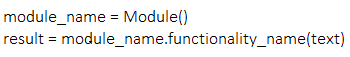}
Table~\ref{tab:standard-flow-features} shows the features supported, output type, and the corresponding functions.

\begin{table*}
\centering
    
\begin{tabular}{|l|l|l|}
\hline
\textbf{Standard Flow module} & \textbf{model\_repo submodule} & \textbf{Huggingface model}\\
\hline
{sentiment} & {SentimentModel} & {\href{https://huggingface.co/l3cube-pune/MarathiSentiment}{MarathiSentiment} \citep{joshi2022l3cube_sent}}  \\ 
{hate} & {HateModel} & 
{\href{https://huggingface.co/l3cube-pune/mahahate-bert}{mahahate-bert} \citep{joshi2022l3cube_hate}} \\
{tagger} & {NERModel} & 
{\href{https://huggingface.co/l3cube-pune/marathi-ner}{marathi-ner} \citep{joshi2022l3cube_ner}} \\
{autocomplete} & {GPTModel} & 
{\href{https://huggingface.co/l3cube-pune/marathi-gpt}{marathi-gpt} \citep{joshi2022l3cube_2}} \\
{mask\_fill} & {MaskFillModel} & 
{\href{https://huggingface.co/l3cube-pune/marathi-bert-v2}{marathi-bert-v2} \citep{joshi2022l3cube_2}} \\
{similarity} & {SimilarityModel} & {\href{https://huggingface.co/l3cube-pune/marathi-sentence-similarity-sbert}{marathi-sentence-similarity-sbert} \citep{joshi2022l3cube_sim}} \\
\hline
\end{tabular}
\caption{\label{modules}Table shows which \textit{model\_repo} submodules are inherited by the Standard Flow modules and the default huggingface model internally used. The huggingface models can be listed and selected in the Model Flow.}
\end{table*}

\subsection{Model Flow}

The \textit{Model Flow} provides advanced functionality intended for use by ML practitioners with knowledge in the NLP domain. It offers flexibility for programmers to select background models and adjust their parameters. The \textit{model\_repo} module defines the model flow. Table~\ref{modules} presents the association between standard flow machine learning-based modules and \textit{model\_repo} submodules.

\section{System Usage}

The mahaNLP library is hosted on the official PyPI repository. It can be installed using the pip command:
\begin{verbatim}
    pip install mahaNLP    
\end{verbatim}
Once installed, we can then import the required features using the python import statement and start utilizing various functionalities. Currently, the library has been tested on x64-bit Windows 10 OS and Google Colab platform (posix-linux).\\

\subsection{Dataset Loading}
A snippet of code demonstrating the loading \textit{mahaSent} dataset provided by mahaNLP is given below:\\
\\
\includegraphics[width=\linewidth]{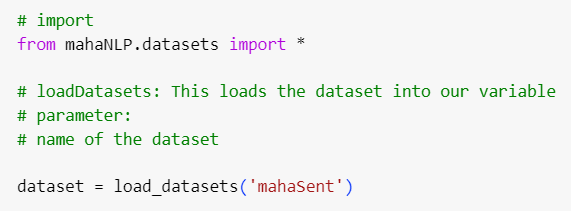}
\subsection{Usage via Standard Flow}
In \textit{Standard Flow}, the user can simply import the feature they want to use (e.g. autocomplete, sentiment, tagger, etc.) and define the object to initialize that particular model. Here, the user can optionally pass the \textit{model\_name} as an argument during model initialization. \\
A simple snippet of generic flow usage for \textit{SentimentAnalyzer} class object is given below:\\
\\
\includegraphics[scale=0.55]{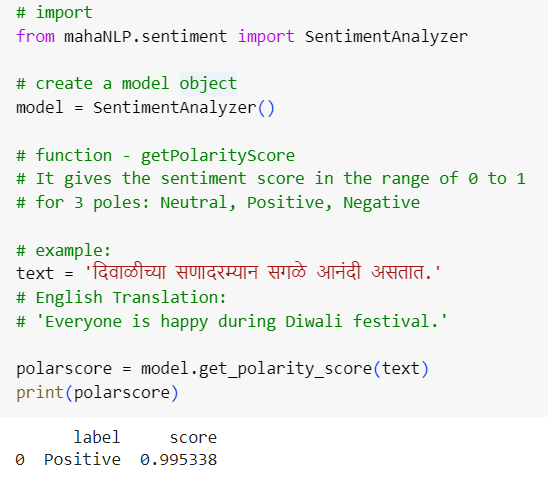} \\
The \textit{get\_polarity\_score} function returns \textit{float} value representing the confidence score for a predicted sentiment class.

\subsection{Usage via Model Flow}
In \textit{Model Flow},  the user has to import the specific model (e.g. mahaHate, mahaNER, etc.) using “import mahanlp.model\_repo.modelname”. Then, the user can define the model object and also can optionally pass the \textit{model\_name} as an argument.
Refer the following code for loading and usage of the \textit{MaskFillModel} class object via \textit{Model Flow}.\\
\\
\includegraphics[scale=0.55]{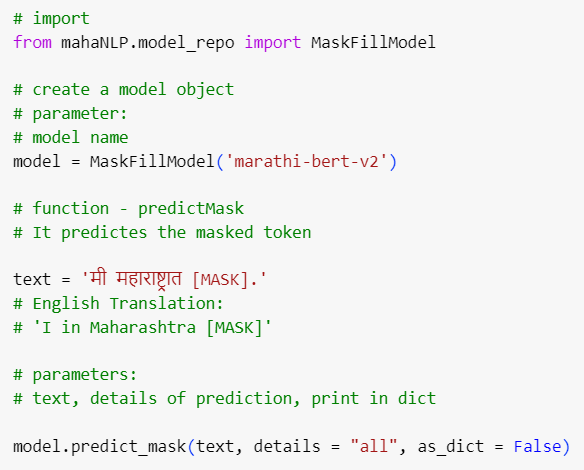}
The predicted token string, sequence, and score can be returned as:\\
\\
\includegraphics[width=\linewidth]{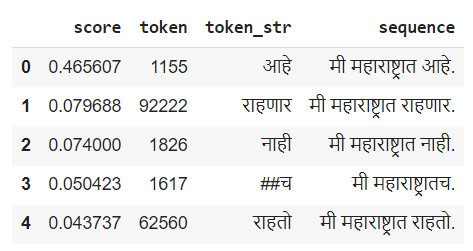}
\\
The library also supports various standard hardware devices. The \textit{gpu\_enabled = True} option allows users to utilize the \textit{}{gpu} for model usage or inference.

Table~\ref{tab:results} demonstrates all the modules of MahaNLP along with inputs and their expected outputs. The detailed description and usage of all the functionalities in mahaNLP are available at the \textbf{\href{https://pypi.org/project/mahaNLP/}{mahaNLP PyPI project}}.




\begin{table*}[ht!]
    \includegraphics[width=\linewidth]{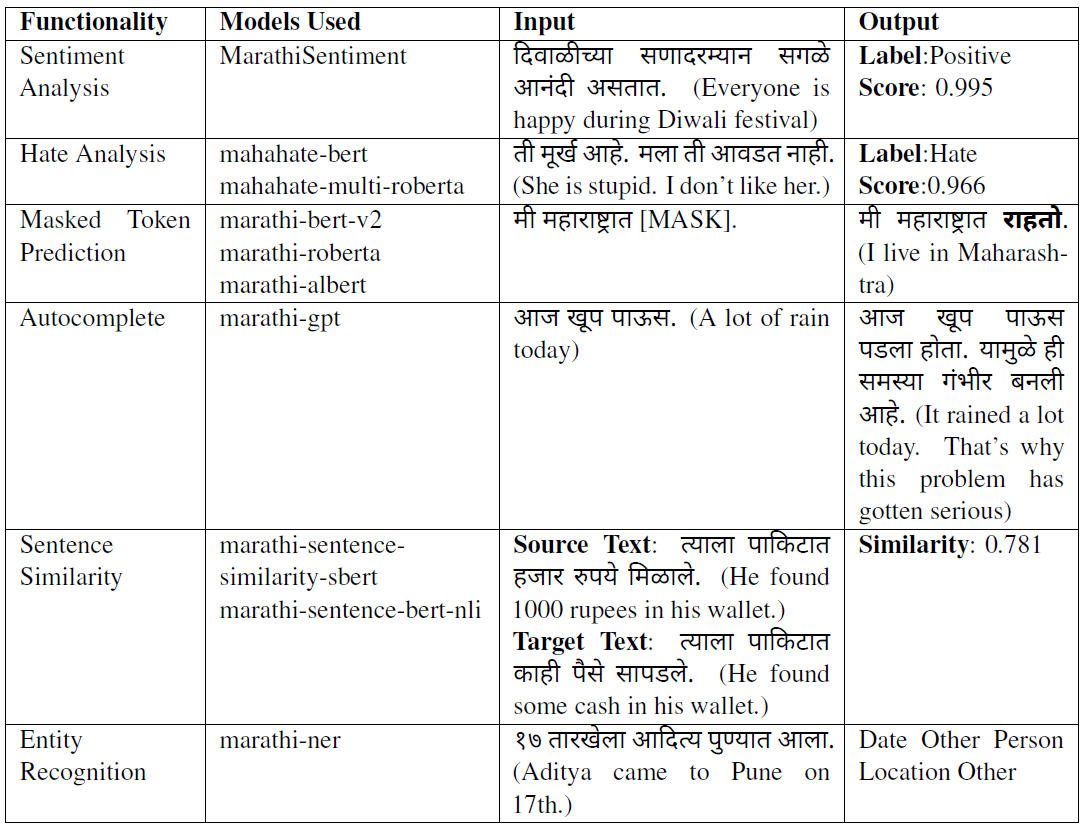}
    \caption{Sample examples of features provided by mahaNLP}
    \label{tab:results}
\end{table*}

\section{Conclusion and Future Work}
We have built mahaNLP - a simple, easy-to-use and extensible toolkit for Marathi language processing and development of robust NLP models. The mahaNLP is majorly built on the top of the huggingface transformers\footnote{\url{https://huggingface.co/l3cube-pune}} library and utilizes pyTorch\footnote{\url{https://pytorch.org/}} framework under the hood. It also utilizes the resources developed by L3Cube as a part of MahaNLP initiative. It primarily provides wrapper classes to perform downstream tasks on supervised datasets, transformer models, and other tools for natural language processing in Marathi. The paper demonstrates the usage of various modules in the mahaNLP library.

We are working on expanding our project scope in the following manner: 
\begin{itemize}
\item Expanding the project to support more low-resource Indic languages.
\item Extending the existing corpora provided by the library and supporting Indo-Aryan language datasets and other regional language datasets like Konkani and Dongari which are very  prevalent in Maharashtra state.
\item Addition of pretrained models for advanced downstream tasks like machine translation, natural language inference etc.
\item Creation of web-based user interactive tools or APIs which can be used by developers at the production level.
\item Further improvements in model accuracy for predictions and reduction in overall model size and computational cost.
\end{itemize}


\section*{Limitations}
\begin{itemize}
    \item Pre-processing can be made more efficient by expanding the dataset for stopwords to a more exhaustive list.
    \item The state-of-art huggingface transformer models have not been tested on a generic domain, but only on specific domains such as the news or social media domain. More generic models built as a part of a broader initiative will be integrated into the library.
    \item The current language models provided by mahaNLP are built on transformers that have high runtime on CPU. In the future, we plan to build and integrate more compact models.
\end{itemize}
These issues are planned to be addressed in the upcoming versions of the library.

\section*{Acknowledgments}
This work was done under the L3Cube Pune mentorship
program. We would like to express our gratitude towards
our mentors at L3Cube for their continuous support and
encouragement. We would also like to thank L3Cube groups CodeBuddies, BitsToBytes, AlgorithmUnlock, and Savadita for their valuable dataset and model contributions.
This work is a part of the L3Cube-MahaNLP project \cite{joshi2022l3cube}.

\bibliographystyle{acl_natbib}
\bibliography{main}
\appendix


\end{document}